\documentclass{article}

\usepackage{arxiv}
\usepackage[utf8]{inputenc} 
\usepackage[T1]{fontenc}    
\usepackage{hyperref}       
\usepackage{url}            
\usepackage{booktabs}       
\usepackage{amsfonts}       
\usepackage{nicefrac}       
\usepackage{microtype}      
\usepackage{graphicx}
\graphicspath{{./figures/}}
\usepackage{amsmath}
\usepackage{tabularx}
\usepackage{subcaption}
\usepackage{float}
\usepackage{comment}
\usepackage{algorithm}
\usepackage{algpseudocode}
\usepackage[numbers,sort&compress]{natbib}

\title{Navigation in a Three-Dimensional Urban Flow using Deep Reinforcement Learning}

\author{
  Federica Tonti \\
  \\Department of Aerospace Engineering, University of Michigan, Ann Arbor, MI 48109, USA \\
  FLOW, Engineering Mechanics, \\
  KTH Royal Institute of Technology, \\
  Osquars backe 18, 11428, Stockholm, Sweden \\
  \texttt{ftonti@umich.edu} \\
  \And
  Ricardo Vinuesa \\
  \\Department of Aerospace Engineering, University of Michigan, Ann Arbor, MI 48109, USA \\
  FLOW, Engineering Mechanics,\\
  KTH Royal Institute of Technology, \\
  Osquars backe 18, 11428, Stockholm, Sweden \\
  \texttt{rvinuesa@umich.edu}
}

\begin{document}
\maketitle

\begin{abstract}Unmanned Aerial Vehicles (UAVs) are increasingly populating urban areas for delivery and surveillance purposes. In this work, we develop an optimal navigation strategy based on Deep Reinforcement Learning. The environment is represented by a three-dimensional high-fidelity simulation of an urban flow, characterized by turbulence and recirculation zones. The algorithm presented here is a flow-aware Proximal Policy Optimization (PPO)  combined with a Gated Transformer eXtra Large (GTrXL) architecture, giving the agent richer information about the turbulent flow field in which it navigates. The results are compared with a PPO+GTrXL without the secondary prediction tasks, a PPO combined with Long Short Term Memory (LSTM) cells and a traditional navigation algorithm. The obtained results show a significant increase in the success rate (SR) and a lower crash rate (CR) compared to a PPO+LSTM, PPO+GTrXL and the classical Zermelo's navigation algorithm, paving the way to a completely reimagined UAV landscape in complex urban environments.
\end{abstract}

\keywords{Deep Reinforcement Learning, UAV, Navigation, Urban environment}

\maketitle

\section*{Introduction}\label{sec1}
The rapidly increasing number of Unmanned Aerial Vehicles (UAVs) in urban environments is due to the number of tasks they can perform, from surveillance and traffic monitoring to package delivery and their ability to reach locations that can be challenging for other aerial vehicles, such as helicopters. This also presents problems related to acoustic pollution and the risk of accidents. It is crucial to develop an efficient strategy for the autonomous navigation of UAVs in complex environments, not only to perform the aforementioned tasks but also to reduce their environmental impact related to acoustic pollution. Path planning becomes essential in order to reduce navigation time and minimize energy consumption. Navigation in urban environments is extremely challenging due to the complexity of the environment itself, with the presence of buildings and complex three-dimensional wind velocity distributions. The velocity field is characterized by the presence of turbulent wakes, vortex shedding, turbulent velocity fluctuations, gusts, recirculation zones and complex interaction phenomena due to the presence of obstacles \cite{Coceal2004,BLOCKEN2015}. \\
Traditional path planning employs deterministic algorithms and heuristic methods. Popular algorithms are potential-field methods \cite{Hwang2002}, grid-based algorithms \cite{Pol2017, SARANYA2014, Champagne2023}, and sampling-based methods such as Rapidly-exploring Random Trees (RRT) \cite{LaValle1998,Zhang2022,Noreen2016,XU2024}, as well as Probabilistic Roadmaps (PRM) \cite{Kavraki1998,bao2025}. Although these algorithms have been shown to be successful in environments which do not exhibit uncertainties, they yield poor performance in dynamic environments. Simultaneously Localization And Mapping (SLAM) algorithms include obstacle detection and avoidance by including a mapping of the environment \cite{Fethi2018,Zixiang2021,Ren2022,Feng2024}, but in large-scale environments they exhibit degraded efficiency since building a map of the whole environment is practically unfeasible.\\
Deep-Learning (DL)-based methods
have recently received significant attention. DL can significantly improve obstacle avoidance and path planning for UAVs by using neural networks to process and interpret sensory data, such as images from cameras, or LiDAR signals \cite{Tang2024, OSCO2021, roghair2021} or ground-based stations \cite{Wang2022}.
These methods allow UAVs to detect and avoid obstacles more
efficiently by recognizing patterns and predicting potential collisions in real time. Deep-Learning methods are extremely
efficient for environments featuring small variations, as they are based on labels and sensitive to environmental
changes. They are suitable for closed geometries but less reliable for urban environments, where the conditions typically change very rapidly. This brings the necessity to develop reinforcement-learning (RL) methods to understand and automate decision-making processes, in which the agent learns by trial-and-error \cite{Sutton2018}. \\
Deep Reinforcement Learning (DRL) has been successfully applied to UAV navigation tasks in complex environments. Bouhamed et al. \cite{Bouhamed2020} used a Deep Deterministic Policy Gradient (DDPG) algorithm to train quadrocopters to navigate in a three-dimensional environment avoiding collisions with static and moving obstacles. Wang et al. \cite{Wang2023} implemented a Faster Regions with Convolutional Neural Network Feature (Faster R-CNN) algorithm for obstacle detection, improved upon a traditional Deep-Q Network (DQN) approach. Collision avoidance capability was significant in unknown or dynamically changing environments. Sheng et al. \cite{Sheng2024} used a Twin Delayed Deep Deterministic Policy Gradient (TD3) to train a UAV to navigate a highly dynamic environment with multiple obstacles achieving high success rates and efficient path planning. Wang et al. \cite{WangJ2024} proposed a Distributed Privileged Reinforcement Learning (DPRL) framework to address partial observability in navigation tasks, outperforming conventional vision-based methods. AlMahamid and Grolinger \cite{AlMahamid2025} introduced Agile DQN, an adaptive deep recurrent attention reinforcement learning method that enhances UAV obstacle avoidance by processing only the most relevant visual regions, achieving faster training and higher performance in simulated 3D environments. Raj and Kos \cite{Raj2024} applied a DQN-based approach for mobile robot navigation in unknown 2D environments, showing effective obstacle avoidance and improved navigation performance. Zhao et al. \cite{Zhao2024} proposed an elastic adaptive DRL strategy to stabilize training in autonomous navigation tasks, demonstrating superior collision avoidance in multi-agent traffic scenarios. Berg et al. \cite{Berg2025} combined deep reinforcement learning with nonlinear model predictive control (NMPC) for autonomous surface vessels, improving digital twin synchronization and obstacle avoidance through adaptive control policies.\
Recently, DRL has been applied to autonomous navigation tasks where the environment is given by high-fidelity simulations. Gunnarson et al. \cite{Gunnarson2021} applied a V-Racer algorithm with a Remember and Forget Experience Replay to discover time-efficient navigation policies to steer a fixed-speed swimmer through unsteady two-dimensional flow fields. Jiao et al. \cite{Jiao2025} showed that gradient sensing was critical for a navigation task in unsteady wake scenarios, applying DRL in a 2D unsteady flow field. The geocentric agent (with access to global direction) could reach the goal by sensing local flow velocity alone. However, the egocentric agent consistently failed with only local velocity sensors. An egocentric navigator needed the extra information on how the flow changed around it to break symmetries and reliably navigate the wake. Tonti et al. \cite{TONTI2025} used a Proximal Policy Optimization (PPO) algorithm \cite{schulman2017proximalpolicyoptimizationalgorithms} in combination with Long Short-Term Memory (LSTM) cells \cite{LSTM1997} for UAV navigation in a two-dimensional urban-like turbulent flow field obtained from a high-fidelity simulation. This study included random start and target locations, as well as random start snapshots, showing the adaptability of the agent in complex flow fields. The present study is an application to a three-dimensional urban environment, represented by a high-fidelity simulation, comparing three different architectures: PPO with LSTM cells, PPO with Gated Transformer XL (GTrXL) \cite{parisotto2019stabilizingtransformersreinforcementlearning}, and PPO with GTrXL with an auxiliary task for flow prediction, in a multi-objective reinforcement learning framework, so that the UAV can further optimize its trajectory exploiting the prediction of the flow field of the next snapshot.\
Proximal policy optimization combined with GTrXL has recently been used for navigation tasks. Huang et al. \cite{Huang2025} applied PPO+GTrXL in an indoor environment with obstacles, comparing the results with a Soft Actor-Critic (SAC) algorithm \cite{2018softactorcriticoffpolicymaximum} combined with a GTrXL. Federici et al. \cite{federici2024meta} applied PPO+GTrXL to a meta-reinforcement learning task applied to spacecraft navigation, showing an improvement in performance with respect to vanilla PPO. \\
To the authors' knowledge, this work is not only the first to bring PPO+GTrXL into a 3D turbulent flow field with obstacles obtained by a high-fidelity simulation, but also the first to integrate a Convolutional Neural Network (CNN) and Gated Recurrent Unit (GRU) encoder, extracting spatial and temporal flow context, into GTrXL, and to train an auxiliary flow prediction head alongside policy and value objectives.  This unified, multi-objective architecture, combining PPO+GTrXL and CNN+GRU embeddings, and transformer-based global attention for both control and real-time flow forecasting, represents a novel contribution to trajectory-optimization tasks.

\section*{Results}\label{results}
In this section, the results of the trained policies are shown and compared.
All episode returns shown below are normalized to a common scale:
\[
R_{\text{norm}}=\frac{R-\min\!\left(R_{\text{all}}\right)}
                      {\max\!\left(R_{\text{all}}\right)-\min\!\left(R_{\text{all}}\right)},
\]
where $R$ is the total reward as defined in Section Methods, $\min(R_{\text{all}})$ and $\max(R_{\text{all}})$ are the
lowest and highest {\em raw} rewards observed over all policies, respectively.  This places
PPO+LSTM, PPO+GTrXL and the proposed Flow-aware GTrXL on the
same $[-1,1]$ range, facilitating a direct comparison of learning speed and
asymptotic performance.\\
Throughout this work, we adopt the following terminology to avoid ambiguity. An environment step refers to a single interaction between the agent and the environment, consisting of selecting an action, receiving a reward, and observing the next state. An episode is a full sequence of steps in the environment from an initial state to a terminal condition. A training iteration corresponds to one optimization cycle, during which the model parameters are updated using gradients computed from batches of collected data. An update step refers specifically to a single gradient descent update within a training iteration, when using several mini-batches per iteration. Performance metrics are plotted against training iterations.\\
The methods compared in this section are described in Methods.

\subsection*{Learning speed and asymptotic return}

\begin{figure}[htbp]
    \centering
    \begin{subfigure}[b]{0.7\textwidth}
        \centering
        \includegraphics[width=\textwidth]{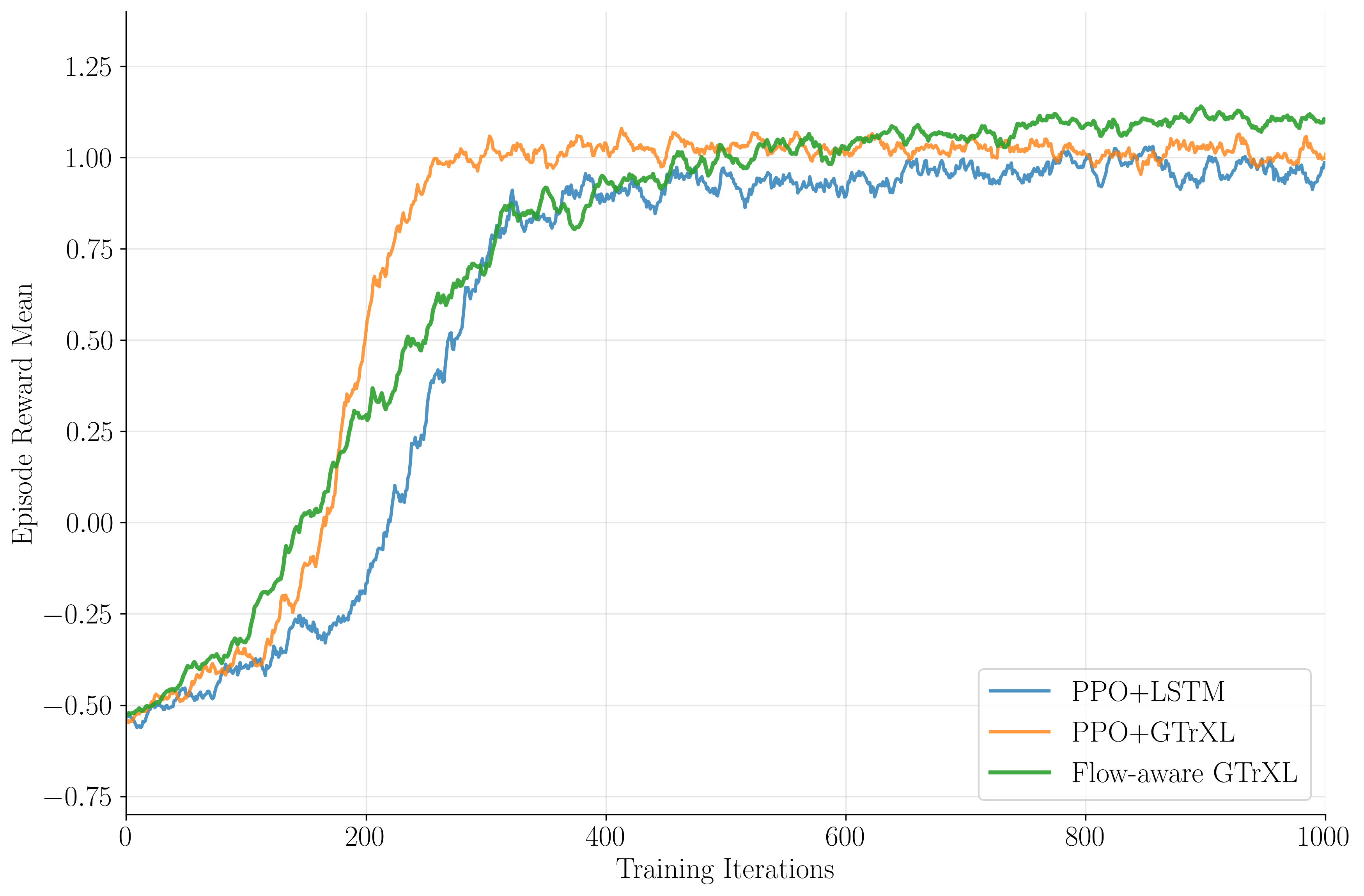}
        \caption{Normalized reward vs. training iterations}
        \label{fig:reward}
    \end{subfigure}
    \hfill
    \begin{subfigure}[b]{0.7\textwidth}
        \centering
        \includegraphics[width=\textwidth]{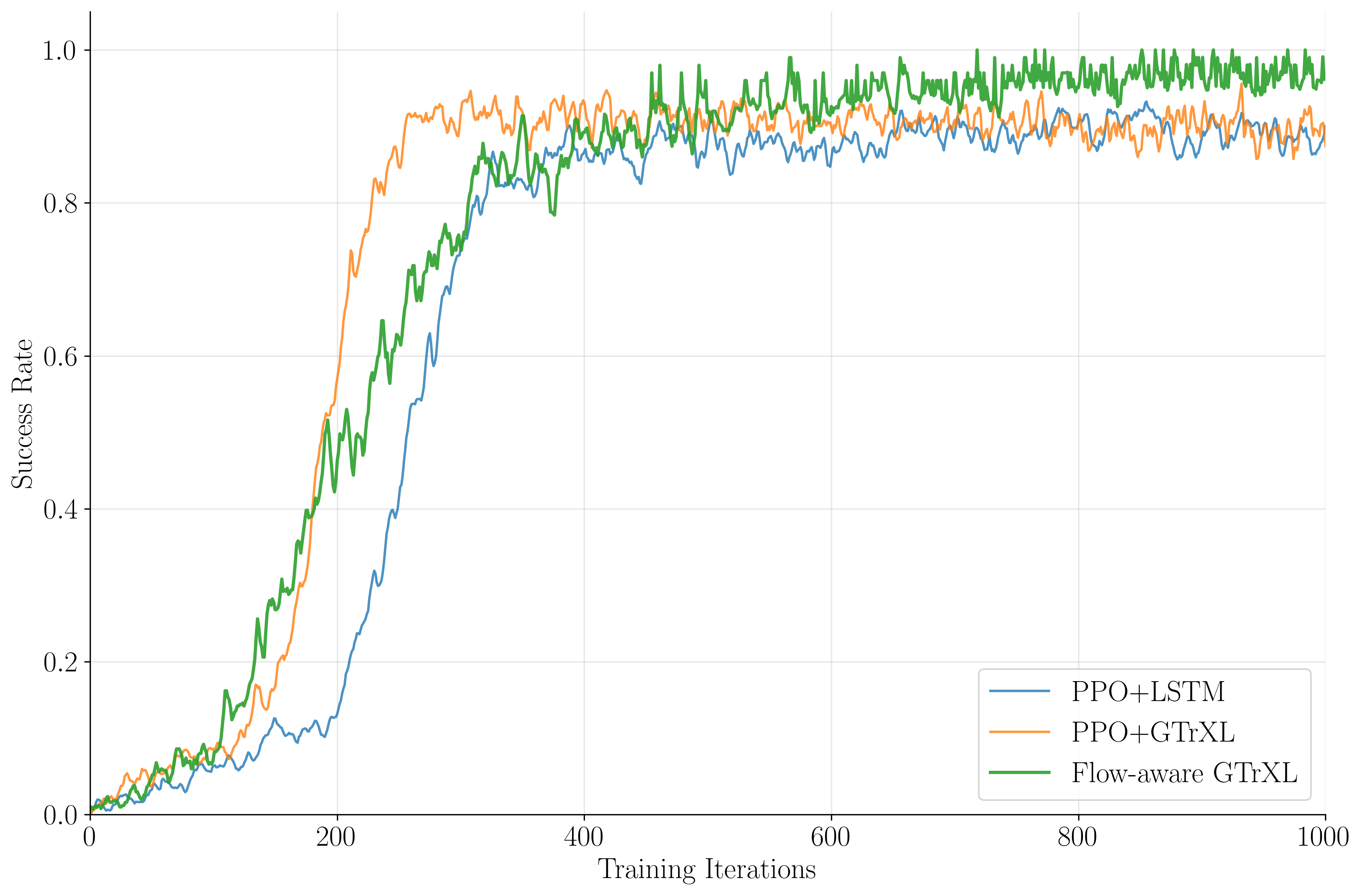}
        \caption{Success rate vs. training iterations}
        \label{fig:success}
    \end{subfigure}
    \hfill
    \begin{subfigure}[b]{0.7\textwidth}
        \centering
        \includegraphics[width=\textwidth]{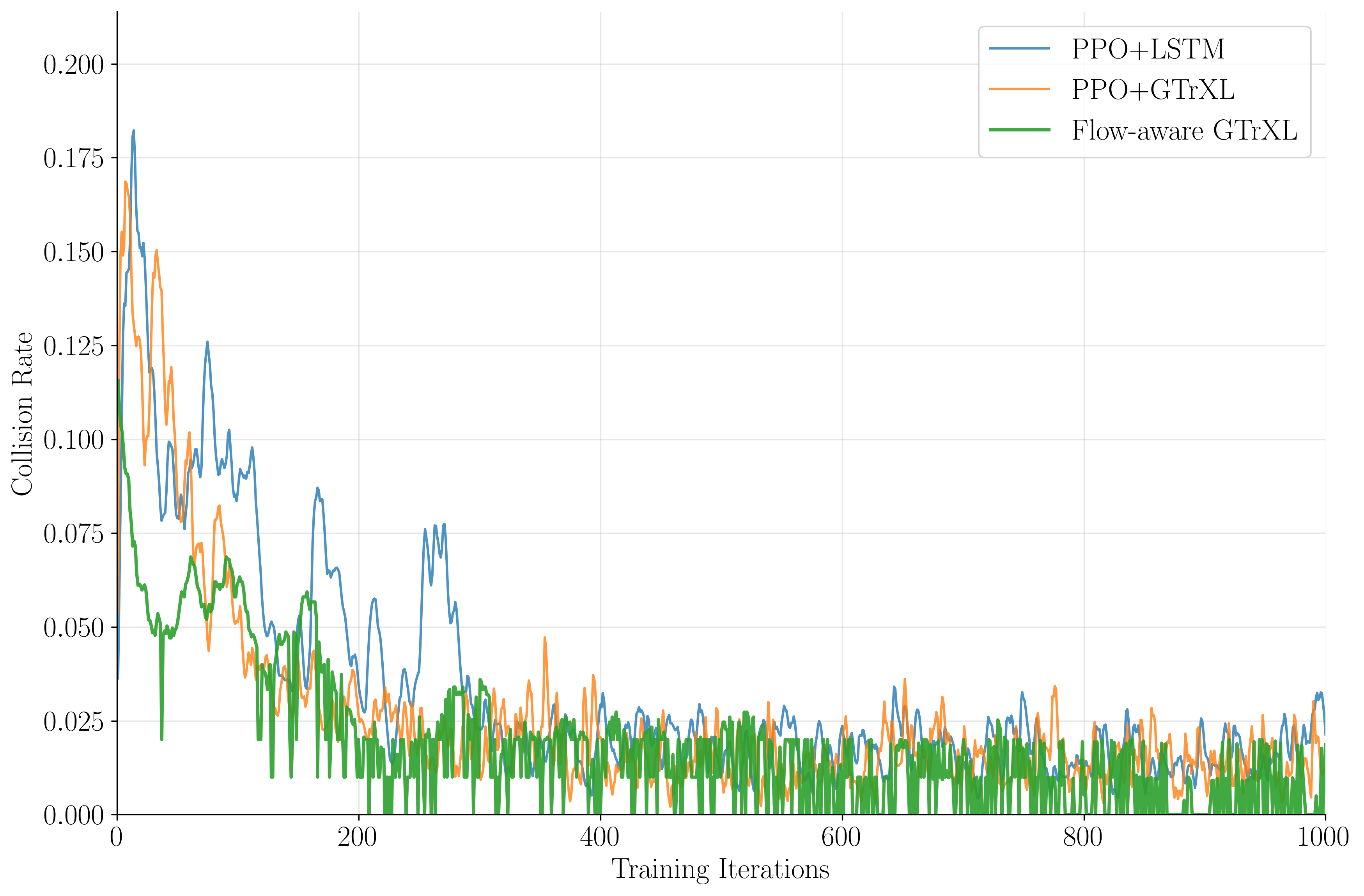}
        \caption{Crash rate vs. training iterations}
        \label{fig:crash}
    \end{subfigure}
    \caption{Evolution of the reward (\ref{fig:reward}), success rate (\ref{fig:success}) and crash rate (\ref{fig:crash}) vs. training iterations for PPO+LSTM (blue), PPO+GTrXL (orange), and Flow-aware PPO+GTrXL (green).}
    \label{fig:metrics}
\end{figure}

Figure~\ref{fig:reward} shows that the LSTM baseline (blue) leaves the
exploration plateau  after 300 training iterations and
saturates at $R_{\text{norm}}\!=\!0.94\pm0.5$.  
Replacing the recurrent core with attention 
raises the plateau to $0.98\pm0.2$.  
Adding the auxiliary flow–prediction head introduces a short
burn-in, yet overtakes the vanilla transformer by iteration~600 and reaches
the top of the scale, $1.0\pm0.1$. The curves are the result of a moving average over 200 episodes.

\subsection*{Task-level safety}

Figure \ref{fig:success} shows that the probability of success during training increases from LSTM to vanilla GTrXL and flow-aware GTrXL, reahcing peaks of 1.0 for the flow-aware method.  
The corresponding crash rates in Figure \ref{fig:crash} drop from PPO+LSTM to Flow-aware GTrXL, confirming that attention reduces the incidence of crashes and
conditioning on flow history drops it further by allowing the
agent to anticipate gust-driven drift, reaching sometimes 0 collisions.\\
During inference, where the algorithms are tested on a unseen environment, we can see that the flow-aware method still reaches a higher performance with respect to the PPO+LSTM and PPO+GTrXL methods.

\subsection*{Classical baseline: Zermelo's optimal navigation}

\begin{table}[htbp]
\label{table_comparison}
\centering
\begin{tabular}{lccccc}
\toprule
\textbf{Metric (mean $\pm$ s.d.)} 
                                  & \textbf{Zermelo}
                                  & \textbf{Flow-GTrXL}
                                  & \textbf{GTrXL}
                                  & \textbf{PPO+LSTM}\\
\midrule
\textbf{Success rate (\%)}       & 61.3      & \textbf{97.6}   & 95.7            & 86.7\\
\textbf{Crash rate (\%)}      & 38.7      & \textbf{0.2}   & 0.4         & 0.5 \\
\bottomrule
\end{tabular}
\caption{Performance of Zermelo's optimization algorithm versus the three RL agents
         on 1\,000 random start–goal pairs}
\end{table}

Although Zermelo’s algorithm computes theoretically optimal trajectories by minimizing a cost function that accounts for control, obstacle avoidance, and final target accuracy, its performance in the unsteady 3D urban environment is significantly inferior to learned policies. Specifically, Zermelo achieves a success rate of only 61.3\%, compared to 97.6\% with flow-aware PPO+GTrXL, 95.7\% with PPO+GTrXL and 86.7\% with PPO+LSTM. This gap is primarily attributed to the open-loop nature of the Zermelo solution: the trajectory is optimized once using a single, randomly selected snapshot of the velocity field and a randomized initial UAV position, without any feedback or re-planning during execution. As a result, the computed trajectory is inherently brittle to perturbations and cannot adapt to the evolving flow field, which is especially detrimental in highly dynamic and partially observable environments.
In contrast, DRL policies are trained in closed-loop settings and are explicitly optimized to generalize across a wide distribution of initial conditions and flow realizations. Architectures like GTrXL and LSTM enable these policies to encode temporal dependencies and respond adaptively to local variations in the flow. Furthermore, DRL agents can implicitly learn obstacle avoidance strategies and exploit transient flow structures through repeated interaction with the environment, capabilities that static optimization-based methods like Zermelo lack. These results demonstrate the critical importance of reactivity and temporal awareness for robust navigation in complex, time-varying environments, highlighting a fundamental limitation of classical optimal control when deployed in real-time.

\begin{figure}[htbp]
  \centering
  \includegraphics[width=0.9\linewidth]{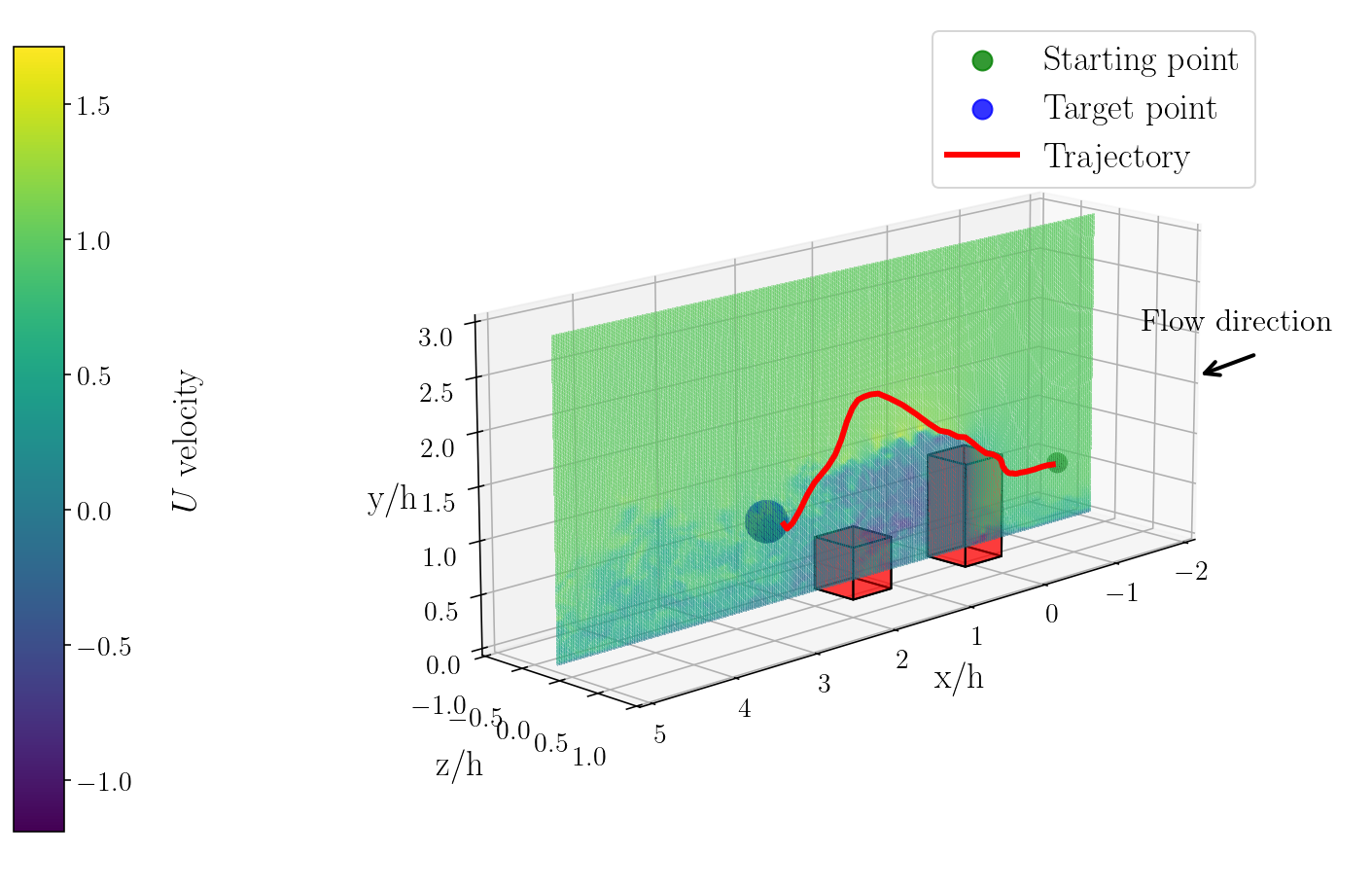}\\[4pt]
  \includegraphics[width=0.9\linewidth]{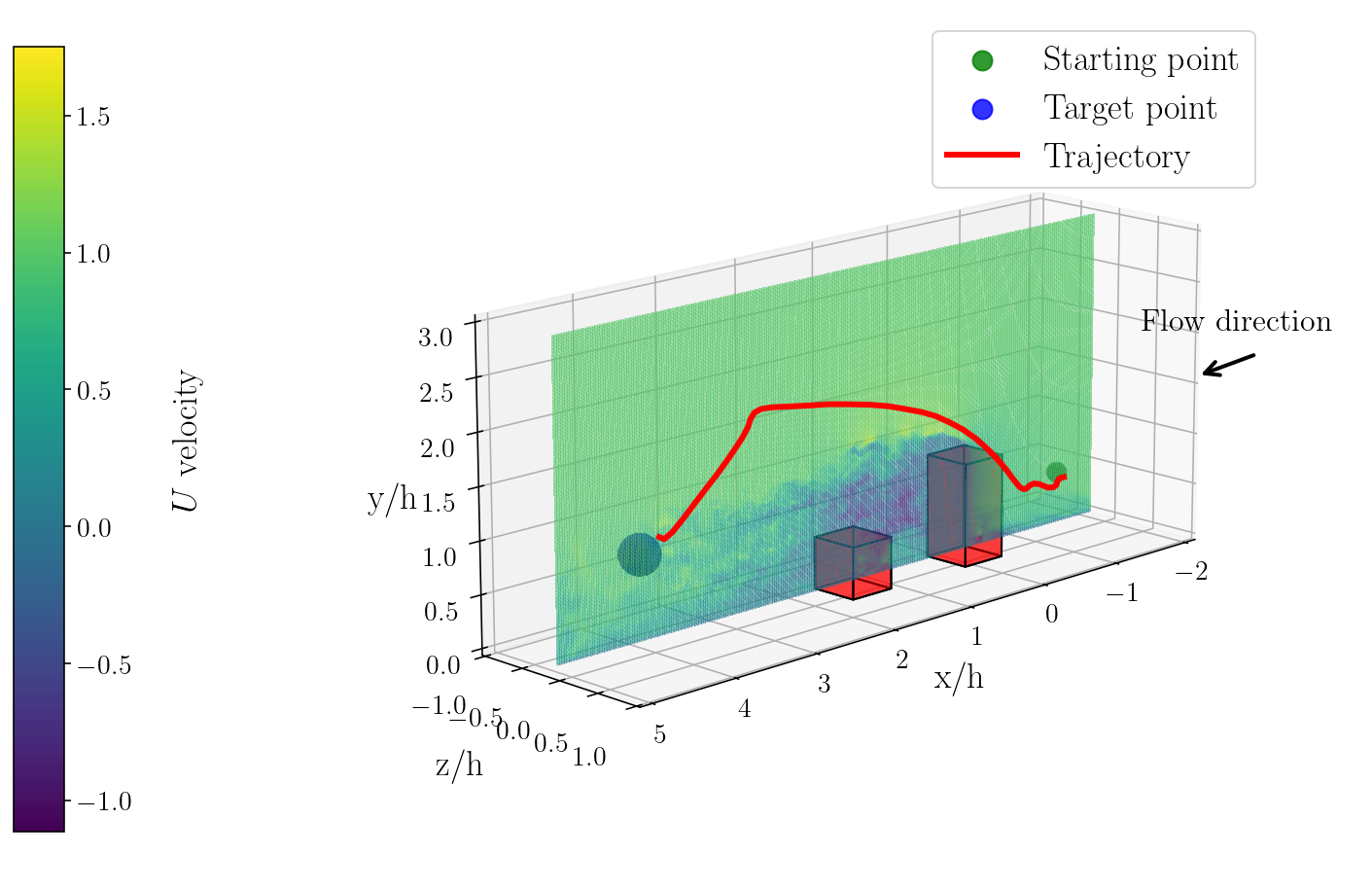}
  \caption{Visualization of trajectories produced by the trained policy of the flow-aware PPO+GTrXL algorithm.}
  \label{fig:traj}
\end{figure}

Figure \ref{fig:traj} shows two examples of trajectories given by the flow-aware PPO+GTrXL. The trajectories are shown in the 3D domain, and a 2D slice of the streamwise velocity field is shown to help the visualization. The environment is represented by a three dimensional high-fidelity simulation of an urban flow. The domain coordinates are $x\in[-2.0,5.0]$, $y\in[0,3.0]$, $z\in[-1.0,1.0]$. The obstacle coordinates are $x_{o1}$$\in[-0.25,0.25]$, $y_{o1}$$\in[0.0,1.0]$, 
$z_{o1}$$\in[-0.25,0.25]$, $x_{o2}$$\in[1.25,1.75]$, $y_{o2}$$\in[0.0,0.5]$, $z_{o2}$$\in[-0.25,0.25]$.  All distances are scaled by the obstacle height $h$. More details can be found in Section Methods.

\section*{Discussion}\label{discussion}
Attention alone can address long‐horizon credit assignment, but at the cost of
a very broad memory pool in which irrelevant observations compete with
critical ones. Conditioning the transformer on a trained flow embedding achieves two complementary effects.  
First, it injects dense stepwise gradients, stabilizing the policy update and
reducing reward variance.  Second, it equips the agent with an implicit
predictive model of the local velocity field, enabling anticipatory actions
rather than purely reactive control.
These mechanisms are supported by the sharp reduction in the collision rate and
the smoother control spectrum observed during the zero-shot evaluation.
\\
The Zermelo's benchmark (Table 1) highlights the
pitfalls of ignoring flow and of open-loop execution: only 61.3\% of success when replayed in the turbulent flow field. In contrast, the flow-aware GTrXL
success rate reaches 97.6\% of the same tasks, without any pre-planning
latency, and produces shorter trajectories.  Even the LSTM baseline
outperforms the Zermelo's benchmark  in reaching the target,
underlining the inadequacy of closed-loop planners in highly disturbed
domains.
Because all returns are normalized to a common $[-1,1]$ range, the curves are directly comparable. While all models exhibit similar initial performance, the Transformer-based architectures (GTrXL and Flow-aware GTrXL) show significantly faster convergence and higher asymptotic reward compared to the LSTM baseline. This is attributed to the architectural differences: LSTMs compress the entire temporal history into a fixed-size hidden state, which limits their ability to capture long-range dependencies, especially in partially observable and dynamically evolving environments. In contrast, GTrXL employs gated self-attention mechanisms with explicit memory access, enabling better retention and exploitation of temporally distant observations, which results in improved policy learning efficiency and stability.
The Flow-aware GTrXL further extends this advantage by incorporating a multi-objective learning framework that includes a contrastive loss aimed at encoding the underlying flow dynamics. This auxiliary task enforces the development of latent representations that are sensitive to changes in the local flow field, thereby enhancing the agent’s ability to react to unsteady flow patterns during navigation. As seen in Figure \ref{fig:metrics}, the Flow-aware GTrXL reaches higher sample efficiency and generalization. These results confirm that coupling policy learning with auxiliary flow representation objectives can significantly improve robustness and adaptivity in complex, time-varying environments.
\\
All experiments were conducted in a high‐fidelity numerical database; real‐world
aerodynamic gusts, sensor noise, and limited onboard computation may
degrade performance.  Furthermore, the auxiliary head uses ground-truth flow
snapshots during training, which is not a possibility during flight. Future work will
replace this with a self-supervised predictor driven by onboard pressure or computer vision tools. Finally, we considered a single UAV; the presence of multiple UAVs may
exhibit interactions that challenge the present architecture.
\\
Our results suggest that 
jointly learning to predict and control is
critical for safe navigation in turbulent 3D environments. Transformer
policies augmented with physically meaningful auxiliary tasks not only match,
but decisively outperform, state-of-the-art sampling planners in robustness and efficiency. We anticipate that similar multi-objective
designs will become standard for autonomous vehicles operating in complex,
partially predictable environments such as urban flow fields or crowded air corridors.

\section*{Methods}\label{methods}
\subsection*{Numerical database and data pre-processing}\label{preprocessing}
The environment is a three-dimensional high-fidelity simulation of an urban turbulent flow field, with two obstacles representing the buildings. The simulation was run using the spectral-element code Nek5000. All the details can be found in Zampino et al. \cite{Zampino2025}. The domain coordinates are $x\in[-2.0,5.0]$, $y\in[0,3.0]$, $z\in[-1.0,1.0]$. The coordinates of the obstacles are $x_{o1}$$\in[-0.25,0.25]$, $y_{o1}$$\in[0.0,1.0]$, 
$z_{o1}$$\in[-0.25,0.25]$, $x_{o2}$$\in[1.25,1.75]$, $y_{o2}$$\in[0.0,0.5]$, $z_{o2}$$\in[-0.25,0.25]$. All distances are scaled by the height of the upstream obstacle $h$. The dataset used for the DRL algorithms is a set of 300 snapshots separated by 0.08750 time units, with the time span of the dataset being 26.25 time units. Time is normalized with $h$ and the freestream velocity $U_{\infty}$. \\
Due to the dimensions of the simulation results, an efficient pre-processing strategy had to be adopted. In order to avoid Input/Output (I/O) overhead during the execution of the algorithm described in subsection Algorithms, the flow field was split into small blocks for each snapshot. The original mesh of the numerical simulation was interpolated on a structured grid of 250$\times$250$\times$250 points. Each snapshot of the simulation has blocks of dimensions 10$\times$10$\times$10 mesh cells in $x,y$ and $z$ grid points. The blocks overlap to ensure that each cell of the domain is covered during the navigation of the UAV, so that the flow field can be continuously mapped. \\
The core of the implementation is the lightweight, scalable framework combined with disk-efficient storage of the decomposed data and a local tricubic interpolation in space and time using KD-tree indexing. 
Algorithm \ref{alg:tricubic_interp} presents a lightweight, on-the-fly extraction of the velocity field in a block-decomposed simulation, performed using tricubic interpolation in space and cubic interpolation in time. Beginning with a mesh file that defines the global domain, the algorithm automatically scans a directory of per-block output files, each named by its time step and spatial indices, to build a time-indexed metadata map. For each stored timestep, it computes the physical spatial bounds of every block, determines the center of each block, and inserts these centers into a KD-tree to enable logarithmic-time nearest-neighbor searches.
At query time for a given position $(x,y,z)$ and time $t$, the algorithm first identifies the bracketing time interval $[t_i, t_{i+1}]$ containing $t$. The it performs spatial interpolation at four control times $\{t_{i-1}, t_i, t_{i+1}, t_{i+2}\}$ to construct a cubic temporal interpolation stencil, with appropriate boundary clamping when fewer than four timesteps are available. For each control time, only the $k$ nearest blocks to the query position are considered, loaded on-demand using an in-memory cache to minimize disk I/O overhead.
Within each relevant block, a parallelized tricubic interpolation routine (implemented via a pybind11 C++ module) computes the velocity components using Catmull-Rom splines over a 4×4×4 neighborhood. When multiple blocks contain the query position, their spatial estimates are combined via inverse-distance weighting to ensure smooth transitions across block boundaries. If no block strictly contains the query point, the algorithm falls back to using the nearest block regardless of its spatial bounds.
Finally, the four spatially-interpolated velocity vectors are blended temporally using Catmull-Rom cubic splines with the interpolation factor $\alpha = (t - t_i)/(t_{i+1} - t_i)$ to produce the final velocity vector $\mathbf{u}(x,y,z,t)$. The caching strategy ensures that once blocks around $(x,y,z)$ are loaded for one control time, they remain available for neighboring times, significantly improving performance in realistic workflows. By never storing the full four-dimensional dataset in memory, this approach scales efficiently to very large simulations while delivering fast and accurate reconstruction in both space and time.

\begin{algorithm}
\footnotesize
\caption{Retrieve velocity $\mathbf{u}(x, y, z, t)$ via tricubic spatial and cubic temporal interpolation}
\label{alg:tricubic_interp}
\begin{algorithmic}[1]
\Require mesh file; directory of block files; sorted time array $T = \{t_0, \ldots, t_{N-1}\}$; query $(x, y, z, t)$; spatial neighbors $k$; cache precision $p$
\Ensure interpolated velocity vector $\mathbf{u}(x, y, z, t)$
\State Load mesh arrays, parse metadata, compute block bounds, build KD-trees
\Function{GetVelocity}{position, $t$, $k$}
  \State position $\gets$ \textsc{ClampPosition}(position); $key \gets$ ($t$, \textsc{QuantizePosition}(position, $p$))
  \If{$key$ in cache} \Return cache[$key$] \EndIf
  \If{$t \leq t_0$} \Return \textsc{SpatialInterpForSnapshot}(position, $t_0$, $k$) \EndIf
  \If{$t \geq t_{N-1}$} \Return \textsc{SpatialInterpForSnapshot}(position, $t_{N-1}$, $k$) \EndIf
  \State Find $i$: $t_i \leq t < t_{i+1}$; Compute $\alpha = ({t - t_i})/({t_{i+1} - t_i})$
  \State Set indices with clamping: $\{i_0, i_1, i_2, i_3\} \gets \{\max(0,i-1), i, i+1, \min(N-1,i+2)\}$
  \For{$j \in \{0,1,2,3\}$} $\mathbf{u}_j \gets$ \textsc{SpatialInterpForSnapshot}(position, $t_{i_j}$, $k$) \EndFor
  \State $\mathbf{u} \gets$ \textsc{CubicTemporalInterp}($\{t_{i_0}, t_{i_1}, t_{i_2}, t_{i_3}\}$, $\{\mathbf{u}_0, \mathbf{u}_1, \mathbf{u}_2, \mathbf{u}_3\}$, $t$, $\alpha$)
  \State cache[$key$] $\gets \mathbf{u}$; \Return $\mathbf{u}$
\EndFunction
\Function{SpatialInterpForSnapshot}{position, timestep, $k$}
  \State Query KD-tree for $k$ nearest blocks to position at given timestep
  \State Initialize list $L \gets \emptyset$
  \For{each block $b_j$ with distance $d_j$}
    \If{position lies within $b_j$'s spatial bounds}
      \State $(u_j, v_j, w_j) \gets$ \textsc{TricubicInterp}($b_j$, position)
      \State Append $(u_j, v_j, w_j, d_j)$ to $L$
    \EndIf
  \EndFor
  \If{$L = \emptyset$} \Comment{Fallback: no blocks contain the query point}
    \State \Return \textsc{TricubicInterp}(nearest block, position)
  \Else \Comment{Multiple blocks contain the point: blend their results}
    \State Compute inverse distance weighted average:
    \State
    \begin{align*}
    \mathbf{u} = \frac{\sum_{(u,v,w,d) \in L} (u,v,w) / (d + 10^{-12})}{\sum_{(u,v,w,d) \in L} 1 / (d + 10^{-12})}
    \end{align*}
    \State \Return $\mathbf{u}$
  \EndIf
\EndFunction
\Function{TricubicInterp}{block\_data, position}
  \State Extract $u, v, w$ grid points; compute bounds; perform tricubic Catmull-Rom on each component
  \State \Return $\mathbf{u}$ = $(u, v, w)$ from tricubic interpolation
\EndFunction
\Function{CubicTemporalInterp}{times, snapshots, $t$, $\alpha$} 
  \State Catmull-Rom cubic spline interpolation using factor $\alpha$
\EndFunction
\end{algorithmic}
\end{algorithm}

During the UAV navigation, the underlying flow field components have to be retrieved to inform the agent with the velocity vector components in real-time.  The procedure is explained in Algorithm \ref{alg:get_flow_tricubic}. First, the position of the UAV is clamped to the valid range of coordinates corresponding to the simulation domain. Then, the query position is quantized with the current time to form a cache key, checking if this specific time-position combination has already been computed. If the cache key is missing, the algorithm proceeds with full tricubic spatial and cubic temporal interpolation.
For temporal interpolation, the algorithm identifies the bracketing time interval and constructs a four-point Catmull-Rom stencil using control times $\{t_{i-1}, t_i, t_{i+1}, t_{i+2}\}$, with appropriate boundary clamping when at the start or end of the time series. At each of these four control times, spatial interpolation is performed using the procedure described in Algorithm \ref{alg:tricubic_interp}, which employs tricubic interpolation within individual blocks and inverse-distance weighting to blend results from multiple overlapping blocks.
The four spatially-interpolated velocity vectors are then combined using Catmull-Rom cubic splines in time with the interpolation factor $\alpha = (t - t_i)/(t_{i+1} - t_i)$. The final result is stored in the cache as the last valid velocity, and the velocity vector is returned. This caching strategy ensures that subsequent queries at nearby positions and times experience minimal computational overhead while maintaining high accuracy in the velocity field extraction.

\begin{algorithm}
\footnotesize
\caption{Retrieve the flow velocity components with full tricubic spatio‐temporal interpolation}\label{alg:get_flow_tricubic}
\begin{algorithmic}[1]
\Require query position $(x,y,z)$; neighbor count $k$; cache precision $p$
\Ensure velocity vector $\mathbf{u}(x,y,z;t)$
\Function{GetVelocity}{position, $t$, $k$}
  \State position $\gets$ \textsc{ClampPosition}(position)
  \State $q_{pos} \gets$ \textsc{QuantizePosition}(position, $p$)
  \State $key \gets$ ($t$, $q_{pos}$)
  \If{$key$ in velocity\_cache}
    \State \Return velocity\_cache[$key$]
  \EndIf
  \If{$t \leq t_0$} \Return \textsc{SpatialInterpForSnapshot}(position, $t_0$, $k$) \EndIf
  \If{$t \geq t_{N-1}$} \Return \textsc{SpatialInterpForSnapshot}(position, $t_{N-1}$, $k$) \EndIf
  \State Find $i$ such that $t_i \leq t < t_{i+1}$
  \State Set indices with clamping: $\{i_0, i_1, i_2, i_3\} \gets \{\max(0,i-1), i, i+1, \min(N-1,i+2)\}$
  \State Define control times $T = \{t_{i_0}, t_{i_1}, t_{i_2}, t_{i_3}\}$
  \State Compute $\alpha = ({t - t_i})/({t_{i+1} - t_i})$
  \For{$j \in \{0, 1, 2, 3\}$}
    \State $\mathbf{u}_j \gets$ \textsc{SpatialInterpForSnapshot}(position, $t_{i_j}$, $k$)
  \EndFor
  \State $\mathbf{u} \gets$ \textsc{CubicTemporalInterp}($\mathbf{u}_{0}, \mathbf{u}_{1}, \mathbf{u}_{2}, \mathbf{u}_{3}; \alpha$)
  \State velocity\_cache[$key$] $\gets \mathbf{u}$
  \State last\_valid\_velocity $\gets \mathbf{u}$
  \State \Return $\mathbf{u}$
\EndFunction
\end{algorithmic}
\end{algorithm}

\subsection*{UAV dynamics}\label{dynamics}
The UAV is modeled as a mass point with six degrees of freedom in translation and two in orientation, in particular yaw $\psi$ and pitch $\vartheta$. The state vector is defined as 

\begin{equation}
\mathbf{s}=[x,y,z,u_g,v_g,w_g,\psi,\vartheta],
\label{eq:UAVstate}
\end{equation}
\\
\noindent and evolves under the combined influence of thrust $V$, the turn rates expressed as $\Delta\psi$ and $\Delta\vartheta$ and the underlying flow field velocity components $\mathbf{u_{flow}}=(u_{f},v_{f},w_{f})$. Note that $u_g,v_g,w_g$ are the $u,v,w$ components of the velocity vector calculated as a sum of the UAV and flow field velocity components in each direction, respectively. Furthermore, $x,y,z$ are the coordinates of the position of the UAV at the current timestep. Since the UAV is represented as a mass point, the roll angle is not included in the system of equations, which is then described as:

\begin{equation}
\label{UAVdynamics}
\begin{cases}
u_{g}= V \cos\vartheta \cos\psi + u_f,\\
v_{g} = V \cos\vartheta \sin\psi + v_f,\\
w_{g}= V \sin\vartheta           + w_f,\\[6pt]
\dot \psi = \dfrac{\Delta \psi}{\Delta t},\\
\\
\dot \vartheta = \dfrac{\Delta \vartheta}{\Delta t}.
\end{cases}\
\end{equation}
\\
\noindent $\Delta\psi$ and $\Delta\vartheta$ are the variations of the yaw and pitch angles in the $\Delta t$ time interval.
\noindent The state of the UAV at time $t+1$ is given by a classical fourth‐order Runge–Kutta integrator (RK4). In this work, $\Delta t$ corresponds to 0.08750 time units and is divided into 40 RK4 substeps. Because the flow field is only updated every $\Delta t$, a single RK4 step would assume that the flow is spatially uniform along the UAV trajectory, leading to significant errors whenever the vehicle traverses regions of strong velocity gradients. By integrating the dynamics at 40 intermediate positions, we faithfully capture these spatial variations. Although RK4 is formally unconditionally stable for smooth ordinary differential equations (ODEs), coupling to a block‐decomposed interpolation grid imposes a pseudo‐Courant–Friedrichs–Lewy (pseudo-CFL) constraint: the UAV must not step more than a small fraction of a block or execute a large angular deflection in one step, otherwise the KD‐tree lookups may skip blocks and cause artifacts. Using multiple RK4 substeps per timestep in the environment ensures a smooth evolution of the state of the UAV. This approach maintains continuity in position and orientation even when the agent applies maximum thrust and angular rate commands. Finally, having 
block‐tree structures and recently accessed block data remaining in memory, the additional integration calls per interval incur negligible extra I/O.  Consequently, the sub‐stepping strategy delivers substantial gains in accuracy and robustness at minimal computational cost. These considerations justify the use of RK4 sub-steps per state update, yielding an effective balance between numerical stability and performance in large‐scale block‐decomposed simulations.

\subsection*{Environment of the DRL algorithm}\label{environment}
\begin{figure}[htbp]
    \centering
    \begin{subfigure}[b]{0.4\textwidth}
        \centering
        \includegraphics[width=\textwidth]{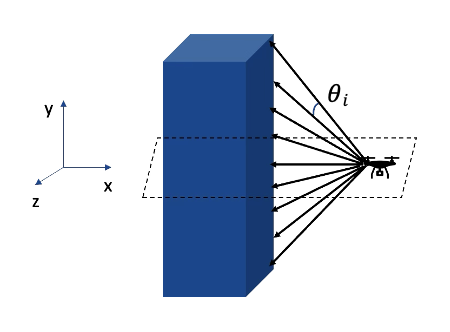}
        \caption{Elevation}
        \label{fig:elevation}
    \end{subfigure}
    \hfill
    \begin{subfigure}[b]{0.4\textwidth}
        \centering
        \includegraphics[width=\textwidth]{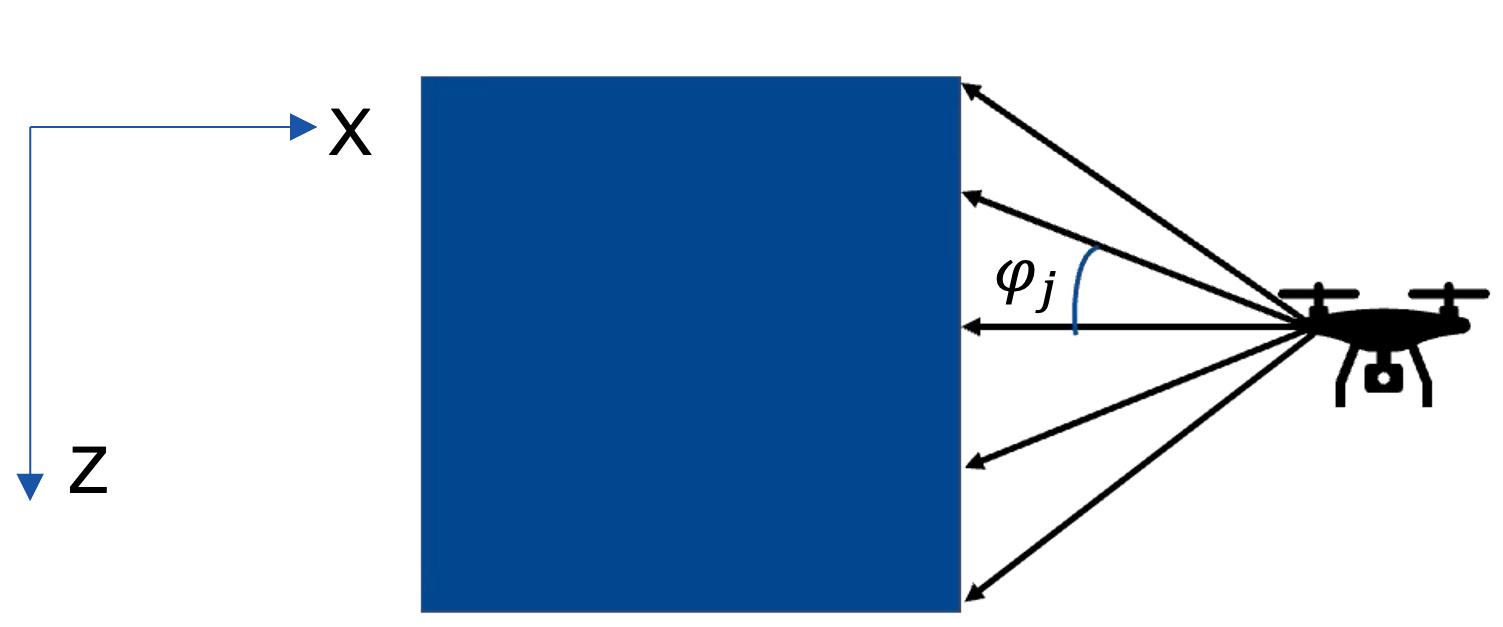}
        \caption{Azimuth}
        \label{fig:azimuth}
    \end{subfigure}
    \begin{subfigure}[b]{0.6\textwidth}
        \centering
        \includegraphics[width=\textwidth]{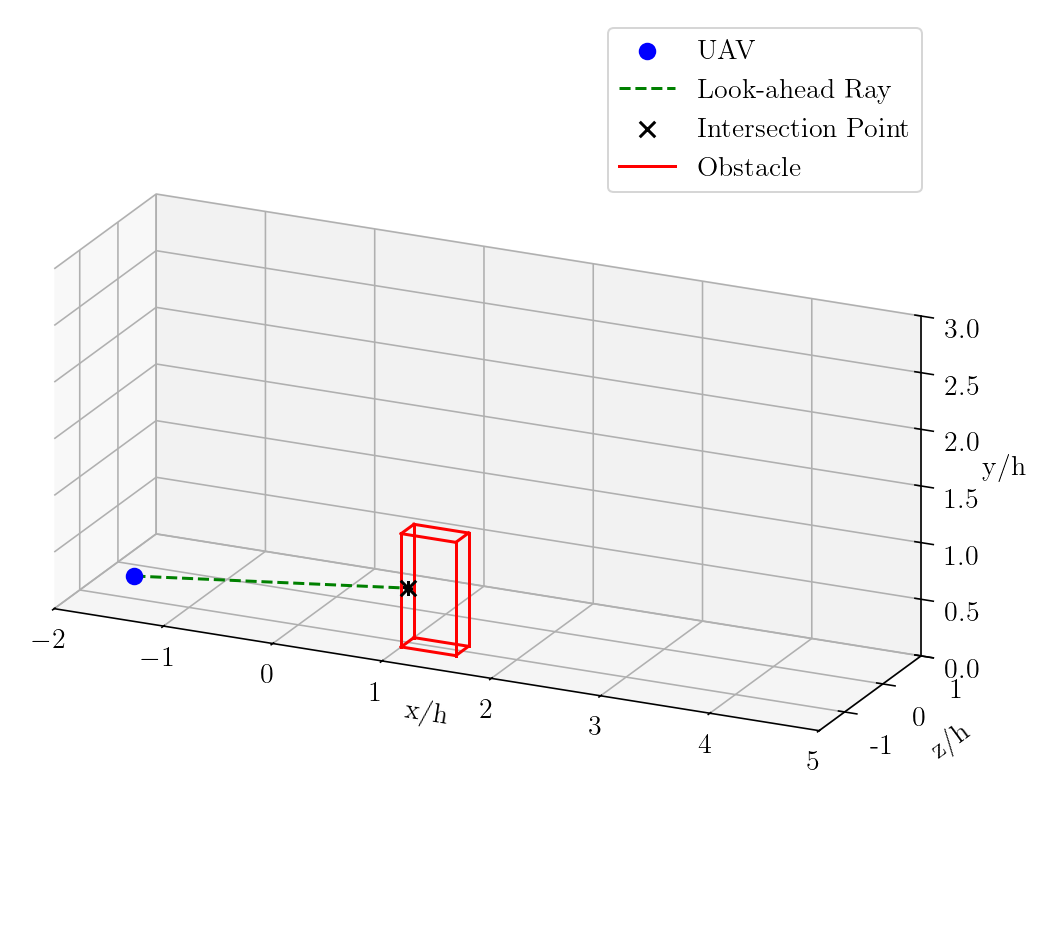}
        \caption{Sketch of the obstacle-detection procedure.}
        \label{fig:obstacledetection}
    \end{subfigure}
    \caption{Sketch of sensors' rays for elevation (\ref{fig:elevation}) and azimuth (\ref{fig:azimuth}). We also show a representation of the obstacle-detection method (\ref{fig:obstacledetection}).}
    \label{elev_azimuth_ray}
\end{figure}

Deep reinforcement learning has been widely used in recent years for control and optimization in fluid mechanics \cite{Guastoni2023,Vignon2023,Font2025}.
The environment is built using Gymnasium \cite{2024gymnasium}, an open-source library for reinforcement learning, focusing on the interface for simulation environments. 
Observation and action spaces are continuous. The observation space is defined as:

\begin{equation}
    \mathbf{o} = \left\{\psi,\vartheta,\psi_{\mathrm{target}}, \vartheta_{\mathrm{target}}, d_{\mathrm{target}}, x,y,z\right\}+\left\{\theta_{i}, \phi_{j}\right\},
    \label{eq:obsspace}
\end{equation}
\\
\noindent where $\psi, \vartheta$ are the yaw and pitch angles of the UAV, $\psi_{\textrm{target}}, \theta_\textrm{{target}}$ the relative yaw and pitch angles of the UAV with respect to the target, $d_\textrm{{target}}$ the Euclidean distance between the UAV and the target, $x,y,z$ the coordinates of the UAV in the domain, and $\theta_{i}, \phi_{j}$ the elevation and azimuth angles associated with the obstacle detection sensors, with $i \in[0,8]$ and $j\in[0,4]$, spanning the angles between $-\pi$ and $\pi$ in their respective directions. Figures \ref{fig:elevation} and \ref{fig:azimuth} graphically show the definition of the environment. 

\noindent The action space is designed to include thrust $V\in[-2.0,2.0]$ and yaw and pitch changes $\Delta\psi \in[-\pi/4,\pi/4]$ and $\Delta \vartheta \in[-\pi/4,\pi/4]$, respectively.\\
Obstacles are detected by providing the agent with a set of directions, since UAVs are in relation to the surroundings typically by images from cameras,
radar signals, or range finders. In this work,  obstacle detection is achieved by implementing a
ray-tracing technique \cite{TONTI2021}. First of all, the UAV has to check for free space in its perspective. The input is the position
of the UAV and the output is a boolean variable which indicates whether the path is free from obstacles or not. Then, if
the obstacle is present, the intersection with the traced rays is computed. First, the direction of the ray is calculated,
based on the ray origin and final point, as well as the coordinates of the obstacles. Then, it is verified whether parallel
directions to the obstacles are present. If the detected directions are not parallel to the obstacles, the intersection point
between the ray and the obstacle is calculated and the distance to the intersection is returned. Figure \ref{fig:obstacledetection} sketches the process. 

\noindent The starting and target points are randomly chosen before the first and after the second obstacle in the domain, as well as the initial orientation of the UAV. The starting snapshot of the algorithm is also randomly chosen among the 300 available. This setup is chosen so that the initial conditions of the flow field themselves exhibit uncertainties. The agent can take a maximum of 100 steps in the domain. \\
The UAV state vector is defined in Equation (\ref{eq:UAVstate}) and it is inferred from the observations, which are given as input to the neural network and described in Equation
(\ref{eq:obsspace}). The problem described here can be then considered as Partially Observable Markov Decision Process (POMDP). A POMDP is characterized by the fact that the agent cannot directly observe the state $s_t$, but receives a set of observations $o_t$ with a distribution $p(o_t|s_t)$. The sequence of observations does not satisfy the Markov property, since $p(o_{t+1}|a_t, o_t, a_{t-1}, o_{t-1}, ..., o_0 )\neq p(o_{t+1}|o_t,a_t)$. Consequently, the agent has to infer the current state $s_t$ based on the history of trajectories.\\

\subsection*{Reward function} \label{reward}
DRL is a process that encourages learning by trial and error and this process is triggered by a reward which is given to the agent when it takes the right actions to complete the assigned task. The structure of the reward is crucial because this guides the agent towards a more effective learning. This component of the algorithm has to be carefully designed and tuned for a specific task. The reward structure has been extended from the 2D version of the problem in Tonti et al. \cite{TONTI2025}. \\
The reward structure is designed to guide the UAV towards the target, while minimizing collisions with obstacles, reducing energy consumption and preventing leaving the designated operational bounds. The reward function is given by several components, each one addressing a different aspect of the UAV's performance. The final reward at each time step \(t\) is the sum of:

\begin{itemize}
  \item \textbf{Transition reward} \(r_{\mathrm{trans}}\): proportional to the reduction in distance to the target,
  \begin{equation}
    r_{\mathrm{trans}} \;=\; \sigma\,d_{\mathrm{dist}}
    \quad\text{where}\quad
    d_{\mathrm{dist}} \;=\; \bigl\lVert \mathbf{x}_{t-1}-\mathbf{x}_{\mathrm{target}}\bigr\rVert
    \;-\;\bigl\lVert \mathbf{x}_{t}-\mathbf{x}_{\mathrm{target}}\bigr\rVert,
  \end{equation}
  where \(\sigma\in\mathbb{R}\) is a scaling constant and $\mathbf{x}$ is the position vector.
  
  \item \textbf{Obstacle penalty} \(r_{\mathrm{obs}}\): an exponential penalty based on the minimum distance \(d_{\min}\) to any obstacle,
  \begin{equation}
    r_{\mathrm{obs}}
    = -\,\xi \,\exp\bigl(-\beta\,d_{\min}\bigr),
  \end{equation}
  with \(\xi,\beta\in\mathbb{R}\), and
  \(
    d_{\min} = \min\{d_1,\dots,d_n\}.
  \)
  
  \item \textbf{Free‐space bonus} \(r_{\mathrm{free}}\):
  \[
    r_{\mathrm{free}}
    = 
    \begin{cases}
      R_{\mathrm{free}}, & \text{if the first‐perspective ray is unobstructed,}\\
      0, & \text{otherwise,}
    \end{cases}
  \]
  where \(R_{\mathrm{free}}\in\mathbb{R}\) is a constant.
  
  \item \textbf{Best‐direction bonus} \(r_{\mathrm{best}}\): when the forward direction is blocked, a small reward in proportion to the chosen changes in angles,
  \begin{equation}
    r_{\mathrm{best}}
    =
    \begin{cases}
      -0.06\,\bigl(\Delta\phi + \Delta\vartheta\bigr),
      & \text{if no free‐space forward,}\\
      0, & \text{otherwise,}
    \end{cases}
  \end{equation}
  where \(\Delta\phi,\Delta\vartheta\) are the yaw and pitch offsets of the best free direction.
  
  \item \textbf{Step penalty} \(r_{\mathrm{step}}\): a constant negative reward per time step,
  \[
    -r_{\mathrm{step}} \in \mathbb{R}.
  \]
  
  \item \textbf{Proximity–velocity penalty} \(r_{\mathrm{prox}}\): discourages high speed when very close to the target,
  \[
    r_{\mathrm{prox}}
    = 
    \begin{cases}
      -0.2\,\bigl\lVert \mathbf{u}_g \bigr\rVert, & \text{if } \lVert \mathbf{x}_t - \mathbf{x}_{\mathrm{target}}\rVert \le 1,\\
      0, & \text{otherwise.}
    \end{cases}
  \]
  
  \item \textbf{Energy penalty} \(r_{\mathrm{energy}}\): proportional to the norm of the propulsion velocity,
  \begin{equation}
    r_{\mathrm{energy}}
    = -\,0.2\,\bigl\lVert \mathbf{u}_g - \mathbf{u}_{\mathrm{flow}}(\mathbf{x}_t)\bigr\rVert.
  \end{equation}
\end{itemize}

Combining all the components, the reward is defined as follows for each time step $t$:
\begin{equation}
  r_t
  = r_{\mathrm{trans}}
  \;+\; r_{\mathrm{obs}}
  \;+\; r_{\mathrm{free}}
  \;+\; r_{\mathrm{best}}
  \;+\; r_{\mathrm{step}}
  \;+\; r_{\mathrm{prox}}
  \;+\; r_{\mathrm{energy}}.
\end{equation}

\noindent At the end of an episode of $m$ $\in\mathbb{N}$ steps, we add:
\begin{itemize}
  \item \emph{Target reached bonus}: if \(\lVert \mathbf{x}_m - \mathbf{x}_{\mathrm{target}}\rVert \le R_{\mathrm{target}}\), then a positive constant is added to the reward and the episode ends. $R_{\mathrm{target}}$ is the radius of the target sphere.
  \item \emph{Collision penalty}: if a collision is detected, a negative constant is added and the episode ends.
  \item \emph{Out‐of‐bounds penalty}: if the UAV exits the domain, \([x_{\min},x_{\max}]\times[y_{\min},y_{\max}]\times[z_{\min},z_{\max}]\),  a negative constant is added, and the episode ends.
  \item \emph{Near‐target bonus}: if \(\bigl|\lVert \mathbf{x}_m - \mathbf{x}_{\mathrm{target}}\rVert - R_{\mathrm{target}}\bigr|<0.5\), then a small positive constant is added.
\end{itemize}

\subsection*{Algorithms}\label{algorithms}
\begin{figure}[h!]
\centering
\begin{subfigure}[b]{0.4\textwidth}
\centering
\includegraphics[width=\textwidth]{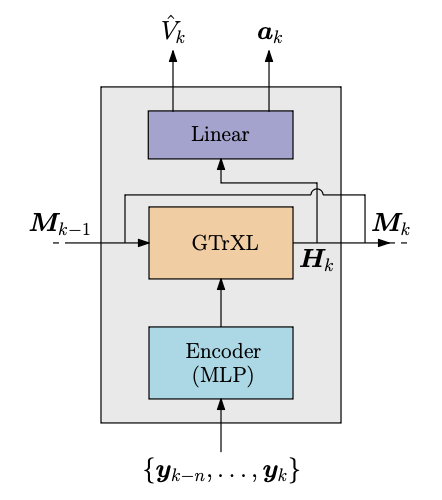}
\caption{PPO+GTrXL.}
\label{fig:ppogtrxl}
\end{subfigure}
\hfill
\begin{subfigure}[b]{0.4\textwidth}
\centering
\includegraphics[width=\textwidth]{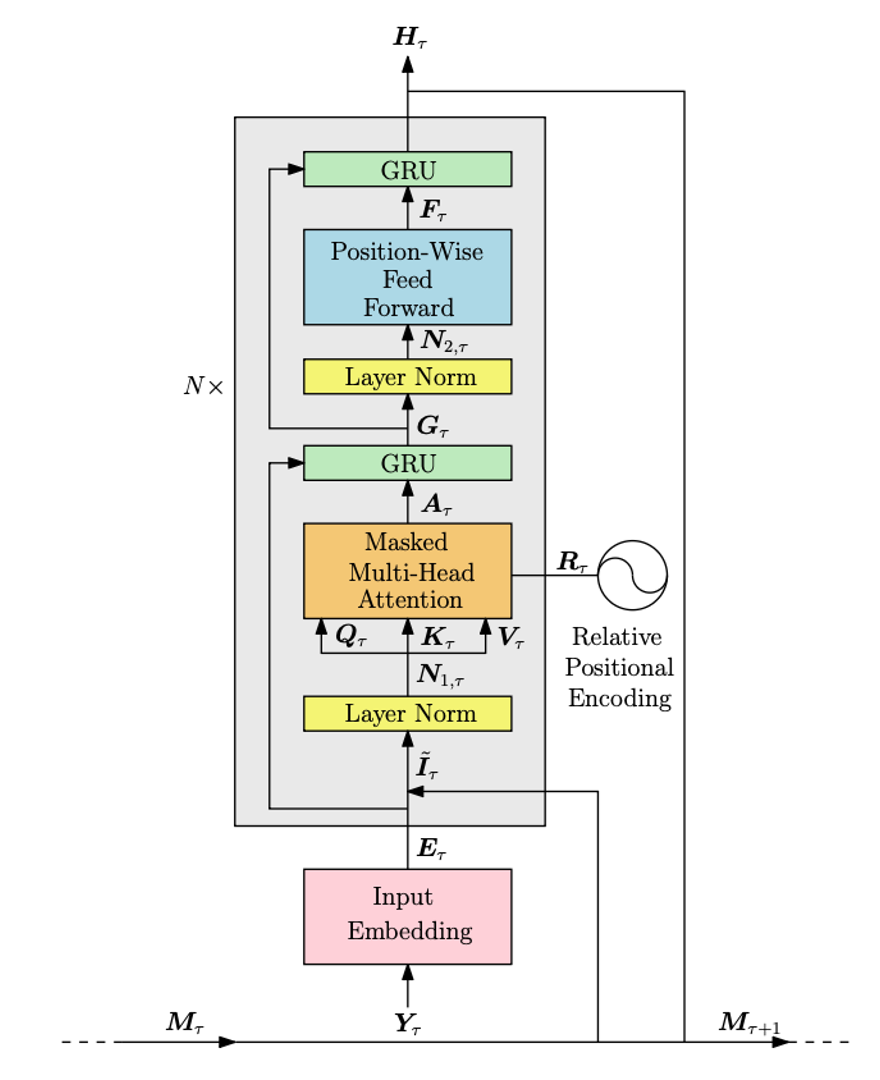} 
\caption{GTrXL block.}
\label{fig:gtrxl}
\end{subfigure}
\hfill
\begin{subfigure}[b]{0.8\textwidth}
\centering
\includegraphics[width=\textwidth]{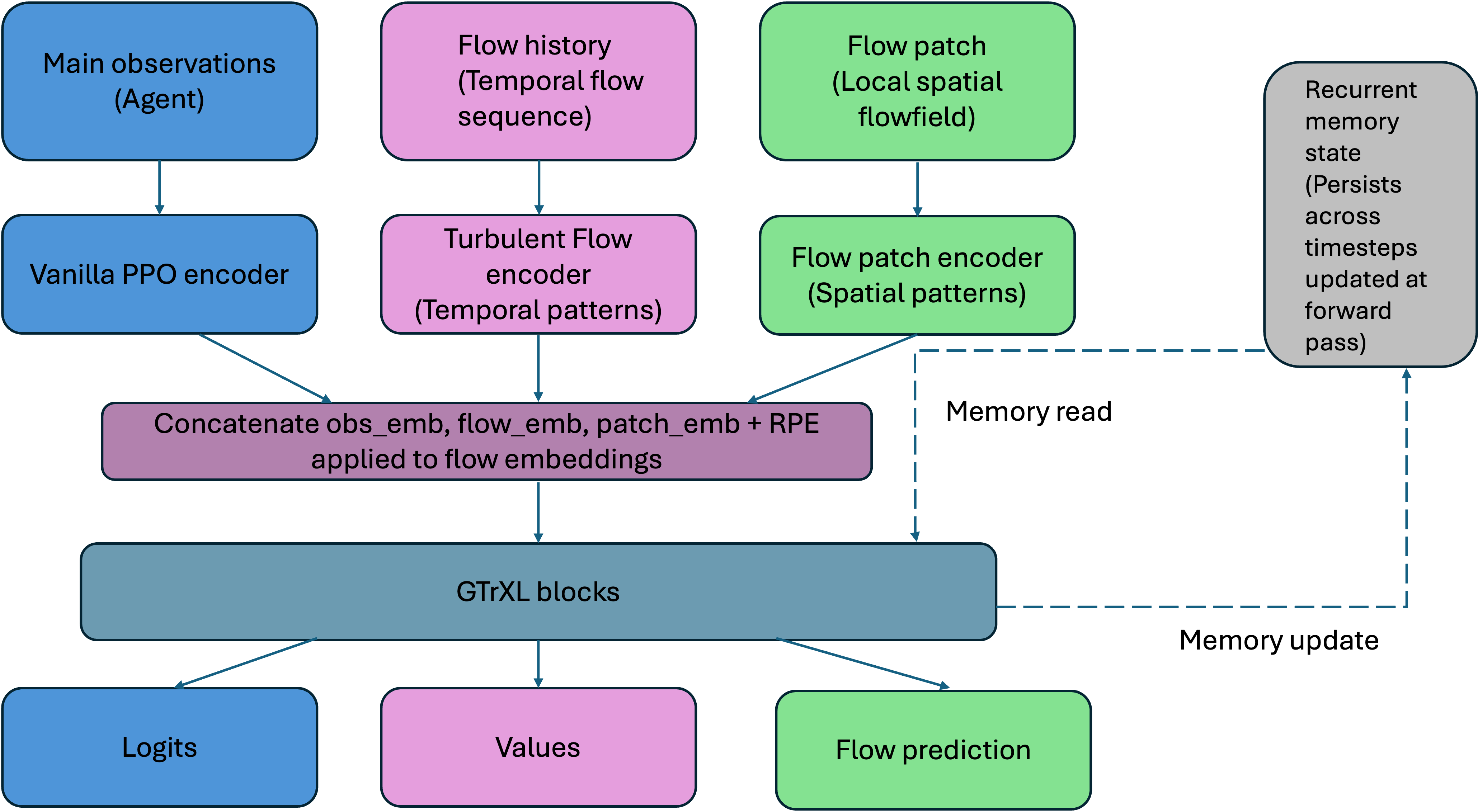} 
\caption{Sketch of the modified algorithm which includes temporal and spatial information of the flow field.}
\label{fig:sketch_algo_cnn}
\end{subfigure}
\label{fig:GTRXL+ppogtrxl}
\caption{PPO+GTrXL (\ref{fig:ppogtrxl}), GTrXL block (\ref{fig:gtrxl}) and sketch of the modified algorithm which includes temporal and spatial information of the flow field (\ref{fig:sketch_algo_cnn})}
\end{figure}

Three different algorithms are compared. The first is the Proximal Policy Optimization (PPO) \cite{schulman2017proximalpolicyoptimizationalgorithms} with Long Short-Term Memory (LSTM) cells \cite{LSTM1997}. The second algorithm tested is a PPO with a Gated Transformer eXtra Large (GTrXL) unit \cite{parisotto2019stabilizingtransformersreinforcementlearning}. The third is a PPO + GTrXL but with a dedicated auxiliary task to integrate a flow prediction head to improve UAV navigation in the environment. The task is then not only trajectory optimization, but a multi-objective DRL where the flow field is also predicted.\\ The training is made on 200 snapshots of the dataset, while 100 snapshots are used for inference to test the learned policies in an unknown environment.\\
In three-dimensional environments, the fixed-size hidden state of the LSTM and the inherently sequential update scheme limit its capacity to maintain and selectively recall the information necessary for effective policy learning. Although LSTM cells can mitigate short-term dependencies via gated memory updates, their recurrence still enforces a strict temporal bottleneck. Each new observation must propagate through all intermediate time steps before influencing the current decision, which can cause important distant events to be “forgotten” or diluted in the hidden state over long horizons. Moreover, LSTMs lack an explicit mechanism for relating non-adjacent states, making it difficult to capture the complex spatial relationships that arise in 3D domains. On the other hand, the GTrXL architecture embeds multihead self-attention layers with gated residual connections, allowing the agent to attend directly to any past observation regardless of its temporal distance. This non-sequential attentional access not only alleviates vanishing-gradient issues but also provides a flexible memory buffer whose capacity grows with trajectory length, enabling more robust encoding of 3D structure and long-range dependencies. Figure \ref{fig:gtrxl} shows the structure of a GTrXL block. The input sequence $Y_{\tau}$ enters the block and first undergoes input embedding ($E_{\tau}$), producing $I_{\tau}$, which is the embedded input. The core processing consists of two sequential components with gated connections. First, the multi-head attention layer computes queries ($Q_{\tau}$), keys ($K_{\tau}$), and values ($V_{\tau}$) from $\tilde{I}_{\tau}$, applies attention with masking to hide future data ($A_{\tau}$) with relative positional encoding ($R_{\tau}$), followed by a GRU gate ($G_{\tau}$) that controls information integration and layer normalization. Second, a position-wise feed-forward network ($F_{\tau}$) is applied, followed by another GRU gate and layer normalization. Both components use skip connections (curved arrows) that enable direct gradient flow. The block maintains recurrent memory states $M_{\tau}$ and $M_{\tau+1}$ that persist across time steps, along with a hidden state $H_{\tau}$ for capturing long-term dependencies. The output consists of the processed sequence and updated memory state $M_{\tau+1}$. The key variables are: $Y_{\tau}$ (input sequence), $E_{\tau}$ (input embedding), $\tilde{I}_{\tau}$ (embedded input), $H_{\tau}$ (recurrent hidden state), $M_{\tau}$ (input memory), $M_{\tau+1}$ (output memory), $Q_{\tau}$/$K_{\tau}$/$V_{\tau}$ (query/key/value matrices), $A_{\tau}$ (attention weights), $R_{\tau}$ (relative positional encoding), $F_{\tau}$ (feed-forward transformation), $G_{\tau}$ (GRU gates). This architecture combines transformer attention mechanisms with RNN gating properties, enabling both effective sequence modeling and stable training dynamics through controlled information flow. Figure \ref{fig:ppogtrxl} shows the PPO+GTrXL architecture, where the MLP encoder is here the PPO algorithm. The observations ${y_{k-n}, ..., y_{k}}$ are the inputs to the encoder, which here is a PPO. The encoded observations are then fed to the GTrXL block, which receives a memory state $M_{k-1}$. The block outputs the hidden state $H_{k}$ and a memory state $M_{k}$. $H_{k}$ is passed to a linear activation layer, which outputs logits $a_{k}$ and values $\hat{V}_{k}$. $M_{k}$ is fed back as input of the GTrXL block.

\noindent The third algorithm used in this work is a modification of the PPO+GTrXL. This custom model implements a PPO agent in which observations and turbulent flow history are first separately encoded, then jointly processed by a stack of GTrXL blocks, and finally decoded into three objectives: action logits, value estimates, and next‐flow‐snapshot predictions. A dedicated encoder ingests a sequence of past flow vectors with a predefined sequence length, applies multiscale 1D convolutions (kernel sizes 3 and 5) to capture short‐ and medium‐term temporal features, fuses them via a linear layer, and feeds the result through a GRU to model longer‐term dynamics, normalizing the output to produce per‐timestep flow embeddings.  In parallel, the agent's proprioceptive characteristics are projected through a PPO in the same embedding space. A small CNN processes a local flow patch into an embedding, which is concatenated along with the flow‐history and observation embeddings.  All tokens are then augmented with learnable relative positional encodings and passed through GTrXL blocks, each combining multi‐head self‐attention with gated residual connections and a feedforward GRU to produce contextually enriched representations.  Finally, the most recent observation token is routed to separate policy and value heads for PPO’s on-policy updates, while the most recent flow token is sent to a flow-prediction head trained with a supervised loss.  By jointly optimizing control and flow forecasting within this multi‐objective architecture, the model leverages transformer-based global attention over CNN+GRU–derived embeddings to navigate and anticipate complex 3D turbulent dynamics in real time. Algorithm \ref{alg:PPO+GTRXL_flow} gives the pseudo-code, and Figure \ref{fig:sketch_algo_cnn} shows a sketch of the algorithm.

\begin{algorithm}
\caption{GTrXL-Enhanced PPO for Turbulent Flow Navigation}
\begin{algorithmic}[1]
\State \textbf{Input:} Raw observations obs, recurrent state state
\State \textbf{Output:} Policy logits, updated state, flow prediction

\State \Comment{Parse multi-modal inputs}
\If{obs is dictionary}
    \State $\text{main\_obs} \gets \text{obs}[\text{``main\_obs''}]$
    \State $\text{flow\_hist} \gets \text{obs}[\text{``flow\_obs''}]$
    \State $\text{flow\_patch} \gets \text{obs}[\text{``flow\_patch''}]$
\Else
    \State Parse flattened tensor based on expected dimensions
\EndIf

\State \Comment{Extract recurrent state}
\State $\text{obs\_buf} \gets \text{state}[0]$ \Comment{Observation buffer}
\State $\text{flow\_buf} \gets \text{state}[1]$ \Comment{Flow buffer}
\State $\text{mems\_in} \gets \text{state}[2:]$ \Comment{Transformer memories}

\State \Comment{Update buffers}
\State $\text{new\_obs\_buf} \gets \text{concat}(\text{obs\_buf}[1:], \text{main\_obs})$
\State $\text{new\_flow\_buf} \gets \text{concat}(\text{flow\_buf}[1:], \text{flow\_hist}[-1])$

\State \Comment{Encode observations}
\State $\text{obs\_emb} \gets \text{obs\_encoder}(\text{new\_obs\_buf})$ \Comment{Standard PPO encoder}
\State $\text{flow\_emb} \gets \text{flow\_encoder}(\text{new\_flow\_buf})$ \Comment{Multi-scale Conv1D + GRU}
\State $\text{patch\_emb} \gets \text{flow\_patch\_encoder}(\text{flow\_patch})$ \Comment{Conv2D + Pool}

\State \Comment{Concatenate embeddings}
\State $x \gets \text{concat}(\text{flow\_emb}, \text{patch\_emb}, \text{obs\_emb})$

\State \Comment{Add relative positional encoding to flow tokens}
\State $x \gets x + \text{RPE}(\text{sequence\_length})$

\State \Comment{GTrXL processing}
\State $\text{mem\_outs} \gets []$
\For{$i = 0$ to $\text{num\_transformer\_units} - 1$}
    \State $x \gets \text{GTrXLBlock}_i(x, \text{memory}=\text{mems\_in}[i])$
    \State $\text{mem\_outs}.\text{append}(x)$
\EndFor

\State \Comment{Extract final representations}
\State $\text{offset} \gets \text{flow\_hist\_len} + 1 + 1$ \Comment{+1 for patch token}
\State $\text{last\_obs} \gets x[:, \text{offset}:][:, -1]$ \Comment{Last observation token}
\State $\text{last\_flow} \gets x[:, :\text{flow\_hist\_len}+1][:, -1]$ \Comment{Last flow token}

\State \Comment{Compute outputs}
\State $\text{policy\_logits} \gets \text{policy\_head}(\text{last\_obs})$
\State $\text{value\_estimate} \gets \text{value\_head}(\text{last\_obs})$
\State $\text{flow\_pred} \gets \text{flow\_pred\_head}(\text{last\_flow})$

\State \Comment{Update recurrent state}
\State $\text{new\_state} \gets [\text{new\_obs\_buf}, \text{new\_flow\_buf}] + \text{mem\_outs}$

\State \Return $\text{policy\_logits}, \text{new\_state}, \text{flow\_pred}$
\end{algorithmic}
\label{alg:PPO+GTRXL_flow}
\end{algorithm}

\noindent The total loss is the sum of the basic vanilla PPO loss and a supervised auxiliary loss on the flow prediction. The auxiliary loss is based on contrastive learning. We propose a contrastive learning framework to improve flow-field representations by learning to distinguish between relevant and irrelevant flow patterns for navigation. Rather than relying solely on reconstruction accuracy, our approach is based on the principle that flows beneficial for navigation should have similar representations, while flows from different contexts should be distinguishable in the learned embedding space.

The contrastive flow loss operates on predicted and target flow fields $\hat{\mathbf{f}}, \mathbf{f} \in \mathbb{R}^d$ by first encoding them through a shared projection network $\phi: \mathbb{R}^d \rightarrow \mathbb{R}^h$ comprising two fully-connected layers with ReLU activation and layer normalization. The resulting embeddings are L2-normalized to unit vectors:
\begin{equation}
\mathbf{e}_{\text{pred}} = \frac{\phi(\hat{\mathbf{f}})}{\|\phi(\hat{\mathbf{f}})\|_2}, \quad \mathbf{e}_{\text{true}} = \frac{\phi(\mathbf{f})}{\|\phi(\mathbf{f})\|_2}.
\end{equation}

We define the positive similarity between predicted and target flows as their scaled dot product:
\begin{equation}
\text{sim}_+ = \frac{\mathbf{e}_{\text{pred}}^T \mathbf{e}_{\text{true}}}{\tau},
\end{equation}
where $\tau > 0$ is a temperature hyperparameter. When negative samples $\{\mathbf{f}_j^-\}_{j=1}^K$ are available (sampled from different temporal steps or spatial locations), we compute negative similarities:
\begin{equation}
\text{sim}_{j}^- = \frac{\mathbf{e}_{\text{pred}}^T \phi(\mathbf{f}_j^-)/\|\phi(\mathbf{f}_j^-)\|_2}{\tau}.
\end{equation}

\noindent The contrastive loss follows the Information Noise Contrastive Estimation (InfoNCE) objective \cite{oord2019representation}, which maximizes mutual information between positive pairs while treating negative samples as noise to contrast against:
\begin{equation}
\mathcal{L}_{\text{contrastive}} = -\log \frac{\exp(\text{sim}_+)}{\exp(\text{sim}_+) + \sum_{j=1}^K \exp(\text{sim}_j^-)}.
\end{equation}

This formulation encourages the model to assign high similarity to the predicted-target pair while maintaining low similarity to negative samples. When negative samples are unavailable, we use the simplified objective $\mathcal{L}_{\text{contrastive}} = -\text{sim}_+$, which directly maximizes  the similarity between predictions and targets. This contrastive formulation encourages the model to learn flow representations that capture meaningful structure, improving the agent's ability to reason about flow dynamics for trajectory planning.
\noindent Finally, the results of the described approaches are compared with a classical optimization algorithm. This work considers the Zermelo's navigation algorithm as applied in \cite{zermelos_algo_2005} ran on the 3D urban environment described in Section Environment of the DRL algorithm. The Zermelo's optimal navigation problem minimizes a cost functional:
\begin{align}
J &= T_f + \int_0^{T_f} \left[ \frac{1}{2} \mathbf{\chi}^\top R \mathbf{\chi} + \phi_{\text{obs}}(\mathbf{x}) \right] \, {\rm dt} + \kappa \left\| \mathbf{x}(T_f) - \mathbf{x}_{\text{target}} \right\|^2, \\
\end{align}
subject to system dynamics, control bounds $|u|$ $\leq$ $u_{max}$, collision avoidance constraints, and domain boundaries. 
The term \( T_f \) represents the final time (mission duration), and its minimization promotes time-efficient flight.
The control effort is penalized by the quadratic term
\[
\frac{1}{2} \mathbf{\chi}^\top R \mathbf{\chi},
\]
where \( \mathbf{\chi} \) is the control input vector (yaw rate, pitch rate, thrust), and \( R \) is a positive definite weighting matrix. This term ensures smooth and energy-efficient maneuvers by discouraging large control inputs.
To avoid collisions, the cost function includes an obstacle avoidance penalty
\[
\phi_{\text{obs}}(\mathbf{x}) = \sum_i \alpha_i \exp(-\beta_i d_i(\mathbf{x})),
\]
where \( d_i(\mathbf{x}) \) is the distance from the current position \( \mathbf{x} \) to the \( i \)-th obstacle, and \( \alpha_i, \beta_i > 0 \) are tuning parameters that control the strength and sharpness of the penalty.
A terminal cost is imposed through
\[
\kappa \left\| \mathbf{x}(T_f) - \mathbf{x}_{\text{target}} \right\|^2,
\]
which penalizes deviations from the desired target position \( \mathbf{x}_{\text{target}} \) at the final time \( T_f \), with \( \kappa > 0 \) controlling the importance of accurately reaching the target.
All cost components involving control and obstacle penalties are accumulated over the trajectory via the integral
\[
\int_0^{T_f} \left[ \frac{1}{2} \mathbf{\chi}^\top R \mathbf{\chi} + \phi_{\text{obs}}(\mathbf{x}) \right] {\rm dt},
\]
which ensures that efficiency and safety are maintained throughout the entire mission duration.
The trajectory is parameterized using B-spline basis functions with control points $C_j$, transforming the infinite-dimensional optimal control problem into a finite-dimensional nonlinear programming problem solved by sequential quadratic programming. Collision constraints are enforced through penalty methods using the environment's geometric obstacle detection, while the flow field influence is incorporated through trilinear interpolation of the discretized velocity data at each trajectory evaluation point.
This comparison is made to enable direct benchmarking between the classical optimization algorithm and DRL-learned policies.

\section*{Acknowledgments}
Federica Tonti and Ricardo Vinuesa acknowledge funding from the European Union’s HORIZON
Research and Innovation Program, project REFMAP, under Grant Agreement number 101096698. The computations were carried out at the supercomputer Dardel at PDC, KTH, and the computer time was provided by the
National Academic Infrastructure for Supercomputing in Sweden (NAISS).

\section*{Data Availability Statement}
All the codes and data used in this work will be made available open access when the article is published here: \url{https://github.com/KTH-FlowAI}

\bibliographystyle{plainnat}
\bibliography{references}

\end{document}